\pdfoutput=1

\documentclass[11pt]{article}

\usepackage{acl}

\usepackage{times}
\usepackage{latexsym}
\usepackage{balance}
\usepackage{booktabs}
\usepackage{multirow}
\usepackage{makecell}
\usepackage{graphicx}
\usepackage{listings}
\usepackage[ruled,vlined]{algorithm2e}
\usepackage{xcolor}

\newcommand{\pluseq}{\mathrel{+}=}
\newcommand{\minuseq}{\mathrel{-}=}
\definecolor{codegreen}{rgb}{0,0.6,0}
\definecolor{codegray}{rgb}{0.5,0.5,0.5}
\definecolor{codepurple}{rgb}{0.58,0,0.82}
\definecolor{backcolour}{rgb}{0.95,0.95,0.92}

\lstdefinestyle{mystyle}{
	backgroundcolor=\color{backcolour},   
	commentstyle=\color{codegreen},
	keywordstyle=\color{magenta},
	numberstyle=\tiny\color{codegray},
	stringstyle=\color{codepurple},
	basicstyle=\ttfamily\footnotesize,
	breakatwhitespace=false,         
	breaklines=true,                 
	captionpos=b,                    
	keepspaces=true,                 
	numbers=left,                    
	numbersep=5pt,                  
	showspaces=false,                
	showstringspaces=false,
	showtabs=false,                  
	tabsize=2
}

\usepackage[T1]{fontenc}

\usepackage[utf8]{inputenc}

\usepackage{microtype}

\title{MIA 2022 Shared Task Submission: Leveraging Entity Representations, Dense-Sparse Hybrids, and Fusion-in-Decoder for Cross-Lingual Question Answering}

\author{Zhucheng Tu and Sarguna Janani Padmanabhan\\
   Apple\\
  \texttt{\{zhucheng\_tu,jananip\}@apple.com}
  }

\begin{document}
\maketitle
\begin{abstract}
We describe our two-stage system for the Multilingual Information Access (MIA) 2022 Shared Task on Cross-Lingual Open-Retrieval Question Answering.
The first stage consists of multilingual passage retrieval with a hybrid dense and sparse retrieval strategy.
The second stage consists of a reader which outputs the answer from the top passages returned by the first stage.
We show the efficacy of using a multilingual language model with entity representations in pretraining, sparse retrieval signals to help dense retrieval, and Fusion-in-Decoder.
On the development set, we obtain 43.46 F1 on XOR-TyDi QA and 21.99 F1 on MKQA, for an average F1 score of 32.73.
On the test set, we obtain 40.93 F1 on  XOR-TyDi QA and 22.29 F1 on MKQA, for an average F1 score of 31.61.
We improve over the official baseline by over 4 F1 points on both the development and test sets.\footnote{Our submission team name is Team Utah: \url{https://eval.ai/challenge/1638/leaderboard/3933}.}
\end{abstract}

\section{Introduction}

This paper describes our submission to the Multilingual Information Access (MIA) 2022 Shared Task on Cross-Lingual Open-Retrieval Question Answering.
Cross-lingual open-retrieval question answering is the task of finding an answer to a knowledge-seeking question in the same language as the question from a collection of documents in many languages.
The answer may not necessarily exist in a document that's in the same language as the question, and hence a system need to find the answer across relevant documents in a different language.
The shared task at Multilingual Information Access 2022 evaluates cross-lingual open-retrieval question answering systems using two datasets, XOR-TyDi QA \cite{asai2020xor} and MKQA \cite{longpre2020mkqa}.\footnote{\url{https://mia-workshop.github.io/shared_task.html}.}

We use a two stage approach, similar to the CORA \cite{asai2021cora} baseline, where the first stage performs multilingual passage retrieval and the second stage performs cross-lingual answer generation.
In the first stage, we leverage mLUKE \cite{ri2021mluke}, a pretrained language model that models entities, to train a dual encoder that encodes the question and passage separately \cite{karpukhin2020dense}.
During retrieval, we perform nearest neighbor search using the query vector on an index of encoded passage vectors.
We merge these dense retrieval hits with BM25 sparse retrieval hits using an algorithm we call Sparse-Corroborate-Dense.
Finally, we feed the ranked list of passages into a reader based on Fusion-in-Decoder \cite{ri2021mluke} and mT5 \cite{xue2020mt5} to produce the final answer.
We do not perform iterative training to repeat these steps multiple times.

Compared to official baseline 1, we improve the macro-averaged score by 4.1 F1 points.
We perform analysis to show the effectiveness of using a multilingual language model with entity representations in pretraining, sparse signals to improve dense hits, and Fusion-in-Decoder.

\section{Data and Processing}

\subsection{Datasets}

We use the official training data consisting of 76635 English questions and answers from Natural Questions  \cite{kwiatkowski2019natural} and 61360 questions and answers from XOR-TyDi QA \cite{asai2020xor} to train our dual encoder model.
We do not train on the development data or the subsets of the Natural Questions and TyDi QA \cite{clark2020tydi} data, which are used to create MKQA or XOR-TyDi QA data.
For training the reader, we leverage Wikipedia language links, which is detailed in Section~\ref{sec:fid}.

XOR-TyDi QA consists of annotated questions and short answers across seven typologically diverse languages.
It can be broken down into two subsets, questions where the answer can be found in a passage in the same language as the question (``in-language''), which just come from answerable questions in TyDi QA \cite{karpukhin2020dense}, and questions where the answer is unanswerable from a passage in the same language as the question and can only be found in an English passage (``cross-lingual''), which are newly added answers in XOR-TyDi QA.
A system should be able to succeed at both monolingual retrieval and cross-lingual retrieval.

MKQA \cite{longpre2020mkqa} consists of 10K parallel questions and answers across 26 typologically diverse locales.
The original question is taken from Natural Questions \cite{kwiatkowski2019natural} in English and translated to 25 different locales.
MKQA does not contain any data for training and is only used for evaluation.

\subsection{Passage Corpus}

We directly use the passages corpus provided by the shared task, with the addition of Tamil (\texttt{ta}) and Tagalog (\texttt{tl}) which are not included in the baseline's passage data.
Following the other languages, we use the \texttt{20190201} snapshot of the Wikipedia dumps.
We follow the same preprocessing steps as the baseline passages data.\footnote{\url{https://github.com/mia-workshop/MIA-Shared-Task-2022/commits/main/baseline/wikipedia_preprocess/build_dpr_w100_data.py}}
We manually split the data into language-specific files, which are later used to build language-specific dense and sparse indices.
Final passage retrieval results are aggregated among different indices.
The number of passages in each language is shown in Table~\ref{table:passage-statistics}.

\begin{table}[h]
	\centering
	\resizebox{\columnwidth}{!}{
			\begin{tabular}{lrr}
					\toprule
					Language & Passages & \% of Total Passages \\
					\midrule
					Arabic (\texttt{ar}) & 1304828 & 2.83 \\
					Bengali (\texttt{bn}) & 179936 & 0.39 \\
					English (\texttt{en}) & 18003200 & 39.00 \\
					Spanish (\texttt{es}) & 5738484 & 12.43 \\
					Finnish (\texttt{fi}) & 886595 & 1.92 \\
					Japanese (\texttt{ja}) & 5116905 & 11.09 \\
					Khmer (\texttt{km}) & 63037 & 0.14 \\
					Korean (\texttt{ko}) & 638865 & 1.38 \\
					Malaysian (\texttt{ms}) & 397396 & 0.86 \\
					Russian (\texttt{ru}) & 4545634 & 9.85 \\
					Swedish (\texttt{sv}) & 4525695 & 9.81 \\
					Tamil (\texttt{ta}) & 219356 & 0.48 \\
					Telugu (\texttt{te}) & 274230 & 0.59 \\
					Tagalog (\texttt{tl}) & 69228 & 0.15 \\
					Turkish (\texttt{tr}) & 798368 & 1.73 \\
					Chinese (\texttt{zh}) & 3394943 & 7.36 \\
					\midrule
					\textbf{Total} & \textbf{46156700} &\textbf{100.0} \\
					\bottomrule
			\end{tabular}}
	\caption{Number of passages in corpus for each language.}
	\label{table:passage-statistics}
\end{table}

%

\section{System Architecture and Pipeline}

Our system differs from the baseline in three ways.
First, in the passage retrieval step, we replace mBERT \cite{devlin2019bert} with mLUKE \cite{ri2021mluke}.
Second, we construct sparse indices from which we will retrieve passages to augment dense retriever-retrieved passages, inspired by \citet{mrtydi} but uses a different dense-sparse hybrid approach.
Finally, we encode each question and passage independently as opposed to all passages together following the Fusion-in-Decoder \cite{izacard2020leveraging} approach.


%

\subsection{mLUKE Dense Retriever}

For dense retrieval, we initialize the dual encoder in DPR \cite{karpukhin2020dense} with a multilingual language model pretrained with entity information, mLUKE \cite{ri2021mluke}, from Wikipedia dumps in 24 languages.
We use the same training objective as DPR and also use the last layer's hidden state of the first input token as the representation for both the question and passage.
Only in-batch negatives are used to train the dual encoder.
mLUKE is pretrained using both the masked language modeling (MLM) task \cite{vaswani2017attention, devlin2019bert} and the masked entity prediction (MEP) task  \cite{yamada-etal-2020-luke}.
Note that entity representations are only used in pretraining input and only word inputs are used in finetuning the dual encoder and inference for simplicity.
This resembles the mLUKE-W variant in the mLUKE paper \cite{ri2021mluke}, which still observed notable improvements over the baselines from only having MEP as an auxiliary task in pretraining.
We use the Hugging Face transformers \cite{wolf-etal-2020-transformers} versions of \texttt{mluke-base} and \texttt{bert-base-multilingual-uncased}.
We build off the official DPR codebase to add the mLUKE encoder.\footnote{\url{https://github.com/facebookresearch/DPR}}

\subsection{Dense-Sparse Hybrids}

\begin{figure}[htp]
	\centering
	\includegraphics[width=1\linewidth]{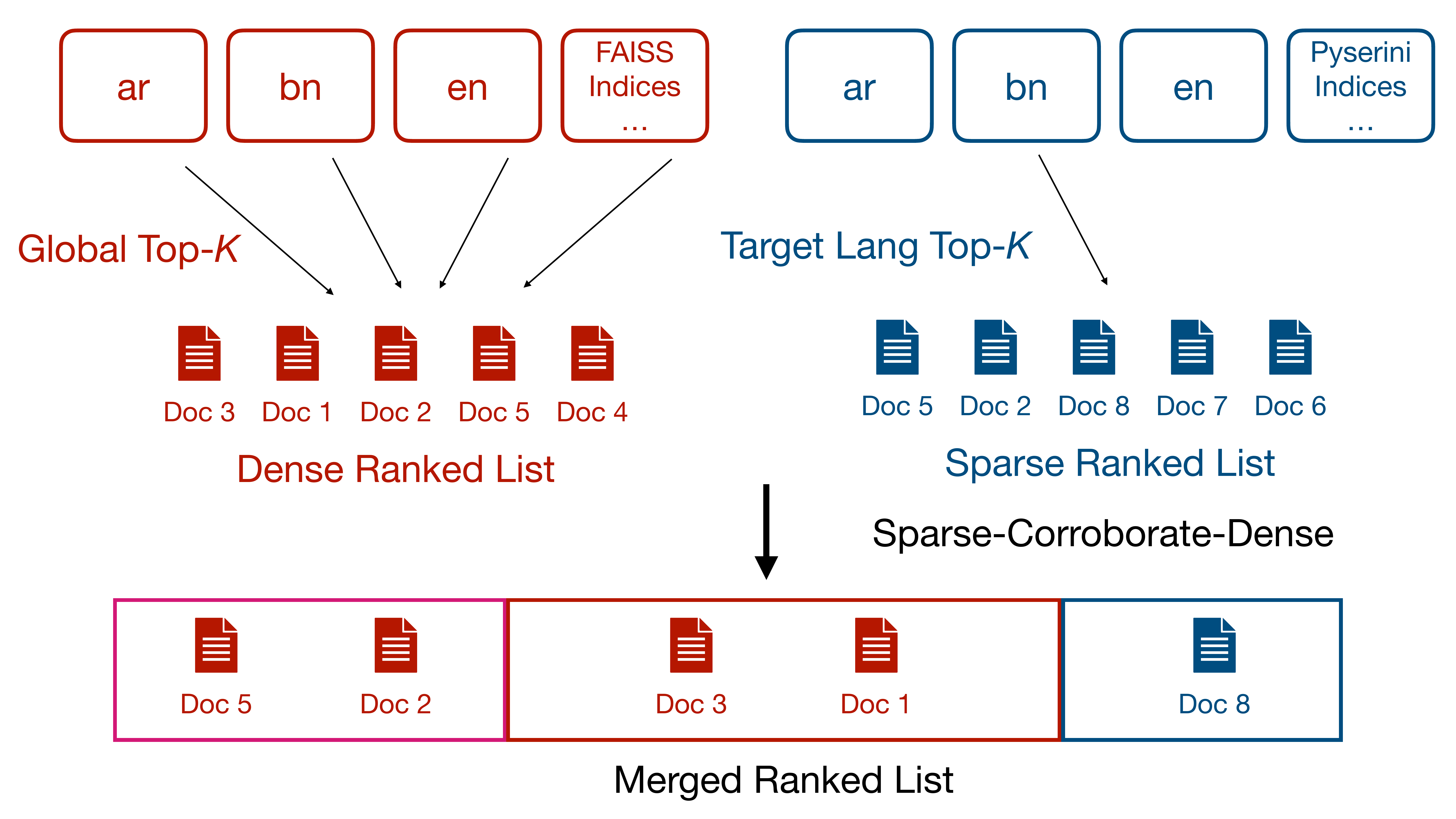}
	\caption{An illustration of the Sparse-Dense-Corroborate algorithm, running with $K=5$  and $max\_frac=0.6$ for a Bengali (\texttt{bn}) question.
	We first retrieve the 5 passages with the highest scores from the dense indices, and the top 5 passages from the Bengali sparse index.
	For the first output slice, we take passages in both lists, ordered by same order as in the dense list, which are doc 5 and doc 2.
	For the second slice, we add the top remaining passages from the dense list, which are doc 3 and doc 1.
	For the third slice, we take the top remaining passages from the sparse list, which is doc 8.
	The max number of sparse results that is allowed to influence the final list is $0.6 \times 5 = 3$, which are docs 5, 2, and 8.
}
	\label{fig:sparse-dense}
\end{figure}

In order to effectively retrieve passages in a multilingual setting, the retrieval component needs to do well in both monolingual retrieval and cross-lingual retrieval.
Monolingual retrieval is the setting where we want to retrieve passages in the same language as the question.
For more than half of the questions in the XOR-TyDi QA dataset, for example, the answer is found in a passage that's in the same language as the question.
Cross-lingual retrieval is the setting where we want to retrieve relevant passages in different language from the question.
We use both sparse retrieval (i.e. BM25) and dense retrieval together in our system.
Experiments in Mr. TyDi \cite{mrtydi} indicate BM25 outperforms DPR \cite{karpukhin2020dense} for the languages in XOR-TyDi QA in the monolingual retrieval setting, but combining the sparse score and DPR score in sparse-dense hybrids perform even better.
At the same time, sparse retrieval rely on lexical token matches and cannot do cross-lingual retrieval effectively without translating the query to the same language as the passage.
To remove the need to use a machine translation system for simplicity, we rely on multilingual dense passage retrieval for cross-lingual retrieval.

For dense retrieval, we use FAISS \cite{johnson2019billion} with \texttt{IndexFlatIP}.
For sparse retrieval, we use Pyserini \cite{yang2017anserini,lin2021pyserini} with BM25 with default parameters.
We build separate indices for each language for both the dense and sparse setting.
For each query, where we want to return $K$ passage, we search for the top $K$ passages globally in the dense indices in all languages, and search for the top $K$ passages in the sparse index in the same language as the question.

We combine results from dense retrieval and sparse retrieval using the following algorithm, which we call Sparse-Corroborate-Dense.
Our final ranked list consists of three ordered slices.
The first slice consists of passages that are present in both dense and sparse retrieved lists, ranked in the same order as they appear in dense retrieval.
The second slice consists of passages that are only in the dense ranked list and not in the sparse ranked list.
The last slice consists of top passages in the sparse ranked list.
The number of passages from the sparse hits that are allowed to influence the final ranked list is no more than $\lfloor max\_frac * K \rfloor$.
We find this works better than the score normalization and combining approach in Mr. TyDi \cite{mrtydi} for cross-lingual retrieval.
Figure~\ref{fig:sparse-dense} has an illustration of this algorithm running with $K=5$ and $max\_frac=0.6$ for a Bengali (\texttt{bn}) question.
Please refer to Algorithm~\ref{algo:sparse-corroborate-dense} for code of the algorithm.

\subsection{Reader}
\label{sec:fid}

Instead of concatenating the question and all the passages in the input to the encoder like in the baseline, which we will call Fusion-in-Encoder, we use the Fusion-in-Decoder (FiD) approach \cite{izacard2020leveraging}.
In Fusion-in-Decoder, the encoder processes each of the \textit{ctxs} passages independently adding special tokens \textit{question:} \textit{lang:} \textit{title:} and \textit{context:} before the question, title and text of each passage, while the decoder performs attention over the concatenation of the resulting representations of all the retrieved passages.
We use mT5 \cite{xue2020mt5} as the encoder-decoder architecture, specially using the \texttt{mt5-base} variant in Hugging Face transformers \cite{wolf-etal-2020-transformers} using the Fusion-in-Decoder codebase.\footnote{\url{https://github.com/facebookresearch/FiD}}

Independent processing of passages in the encoder allows to scale linearly to large number of contexts, while processing passages jointly in the decoder helps better aggregate evidence from multiple passages.

In order to semantically ground the entities across different languages together, we use Wikipedia language links to augment the data from retriever while training FiD based reader, like the CORA baseline.
First, for each question in the MIA training set that comes from Natural Questions, we use the answer to search for the corresponding Wikipedia page using the Wikipedia API.
Generally, only answers that are entities will have a result.
This returns the titles of the Wikipedia articles in different languages, which we use as the answer in different languages.
We use the DPR checkpoint trained with adversarial examples to retrieve English passages from the index.\footnote{\url{https://github.com/facebookresearch/DPR\#new-march-2021-retrieval-model}}
For each English question-answer pair, we find corresponding entries in other languages being evaluated in the task and generate $[Query_{eng}, Lang_{target}, Answer_{target}, Passages]$ tuples for training FiD.
This data is augmented to the original training data provided by the retriever.

\section{Results}

For training the dual encoder, we use the official training data without any hard negatives with the same hyperparameters as the baseline dual encoder \cite{asai2021cora}.
For training Fusion-in-Decoder, we combine the all of the retrieval results with sampled Wikipedia language link augmented passages such that the total percentage of training examples from either source is 50\%.
For Wikipedia language link augmentation, we only sampled English answers from the training set with links to 10 or more languages.
We use the baseline retrieval results instead of mLUKE-retrieved results to develop the retriever and reader in parallel.
We use learning rate of 0.00005 with linear learning schedule with a weight decay of 0.01 using the AdamW optimizer.
The context size (number of passages) in the final submission is 20 passages.
Note for retrieval we use $K=60$ to use the same retrieval results for different context size experiments, but in the final submitted system take the top 20 from this list for the reader.
We use $max\_frac=0.2$ for Sparse-Corroborate-Dense.
We use the best checkpoint on the development set for both components.

\subsection{Main Results}

\begin{table*}[h]
	\centering
	\resizebox{\linewidth}{!}{
		\begin{tabular}{lcccccccc|ccccccccccccc}
			\toprule
			& \multicolumn{8}{c}{XOR-TyDi QA F1} & \multicolumn{13}{c}{MKQA F1} \\
			System & \texttt{ar} &  \texttt{bn} &  \texttt{fi} &  \texttt{ja} &  \texttt{ko} &  \texttt{ru} &  \texttt{te} & Avg & \texttt{ar} & \texttt{en} & \texttt{es} & \texttt{fi} & \texttt{ko} & \texttt{ms} & \texttt{ja} & \texttt{km} & \texttt{ru} & \texttt{sv} & \texttt{tr} & \texttt{zh\_cn} & Avg  \\
			\midrule
			Baseline 1 & 51.29 & 28.72 & 44.35 & 43.21 & 29.84 & 40.68 & 40.19 & 39.76 & 8.77 & 27.86 & 24.92 & 23.25 & 8.28 & 22.64 & 15.18 & \textbf{5.73} & 14.00 & 24.13 & 20.60 & 13.14 & 17.38  \\
			\textbf{Our submission} & \textbf{54.84} & \textbf{30.68} & \textbf{47.41} & \textbf{47.29} & \textbf{33.90} & \textbf{43.13} & \textbf{47.00} & \textbf{43.46} & \textbf{13.34} & \textbf{39.57} & \textbf{29.74} & \textbf{24.73} & \textbf{12.14} & \textbf{27.44} & \textbf{18.97} & 2.57 & \textbf{19.36} & \textbf{28.26} & \textbf{25.52} & \textbf{22.29} & \textbf{21.99}  \\
			\bottomrule
	\end{tabular}}
	\caption{End-to-end development set results. Baseline 1 and our submission obtain overall macro-averaged F1 scores of 28.57 and 32.73 respectively. Our submission outperforms the baseline on all languages except Khmer (\texttt{km}) on MKQA.}
	\label{table:overall-dev}
\end{table*}

\begin{table*}[h]
	\centering
\resizebox{\linewidth}{!}{
	\begin{tabular}{lcccccccc|ccccccccccccc|cc}
		\toprule
		& \multicolumn{8}{c}{XOR-TyDi QA F1} & \multicolumn{13}{c}{MKQA F1} & \multicolumn{2}{c}{Sup} \\
		System & \texttt{ar} &  \texttt{bn} &  \texttt{fi} &  \texttt{ja} &  \texttt{ko} &  \texttt{ru} &  \texttt{te} & Avg & \texttt{ar} & \texttt{en} & \texttt{es} & \texttt{fi} & \texttt{ko} & \texttt{ms} & \texttt{ja} & \texttt{km} & \texttt{ru} & \texttt{sv} & \texttt{tr} & \texttt{zh\_cn} & Avg & \texttt{ta} & \texttt{tl}  \\
		\midrule
		Baseline 1 & 49.66 & 33.99 & 39.54 & 39.72 & 25.59 & 40.98 & 36.16 & 37.95 & 9.52 & 36.34 & 27.23 & 22.70 & 7.68 & 25.11 & 15.89 & \textbf{6.00} & 14.60 & 26.69 & 21.66 & 13.78 & 17.14 & 0.00 & 12.78  \\
		\textbf{Our submission} & \textbf{55.33} & \textbf{30.48} & \textbf{41.01} & \textbf{43.45} & \textbf{31.21} & \textbf{42.62} & \textbf{42.40} & \textbf{40.93} & \textbf{12.67} & \textbf{39.63} & \textbf{30.85} & \textbf{25.22} & \textbf{12.18} & \textbf{29.09} & \textbf{20.49} & 2.36 & \textbf{18.82} & \textbf{29.62} & \textbf{26.16} & \textbf{22.60} & \textbf{22.29} & \textbf{20.75} & \textbf{20.95}  \\
		\bottomrule
\end{tabular}}
\caption{End-to-end test set results. Baseline 1 and our submission obtain overall macro-averaged F1 scores of 27.55 and 31.61 respectively. ``Sup'' indicates the surprise languages.  Our submission outperforms the baseline on all languages except Khmer (\texttt{km}) on MKQA.}
	\label{table:overall-test}
\end{table*}

We first report end-to-end results using our best system compared to the baseline in Table~\ref{table:overall-dev} for the development set and Table~\ref{table:overall-test} for the test set.
On the development set, we obtain macro-averaged F1 score of 43.46 across all languages on XOR-TyDi QA, an improvement of 3.70 F1 points over 39.76 obtained by baseline 1.
We obtain macro-averaged F1 score of 21.99 across all languages on MKQA, an improvement of 4.61 F1 points over 17.38 obtained by baseline 1.
On the test set, we observe fairly consistent results compared to the development set.
On XOR-TyDi QA, our system and baseline 1 obtains 40.93 and 37.95 respectively, an improvement of 2.98 F1 points.
On MKQA, our system and baseline 1 obtains 22.29 and 17.14 respectively, an improvement of 5.15 F1 points.
On both the development set and test set, we outperform the baseline on all languages except for Khmer (\texttt{km}) on MKQA.

We observe our system frequently retrieves irrelevant passages for Khmer through qualitatively sampling some passages retrieved for Khmer questions, providing little chance for the reader to find the answer.
mLUKE uses 24 languages for pretraining and does not include Khmer, making it difficult to align entities in Khmer.
Furthermore, even if we use the baseline retrieval results, we still see a large drop in reader effectiveness when we switch to Fusion-in-Decoder from row (iii) to (ii) in Table~\ref{table:ablation-overall-dev}.
We only have 3101 rows in the training data for Khmer for our reader all from Wikipedia language links, out of 275990 rows in total.

On the surprising languages Tagalog (\texttt{tl}) and Tamil (\texttt{ta}), we outperform the baseline by a large margin.
Perhaps surprisingly, this large improvement cannot be attributed to the presence of Tagalog and Tamil passages in our corpus, since in our best submission, for example, out of the 350 Tamil questions, only one question has a retrieved passage in Tamil in the top results that are fed to the reader.
Instead, the system is able to generate correct answers from English passages.

\subsection{Analysis}

\begin{table*}[h]
	\centering
	\resizebox{\linewidth}{!}{
		\begin{tabular}{lcccccccc|ccccccccccccc}
			\toprule
			& \multicolumn{8}{c}{XOR-TyDi QA F1} & \multicolumn{13}{c}{MKQA F1} \\
			System & \texttt{ar} &  \texttt{bn} &  \texttt{fi} &  \texttt{ja} &  \texttt{ko} &  \texttt{ru} &  \texttt{te} & Avg & \texttt{ar} & \texttt{en} & \texttt{es} & \texttt{fi} & \texttt{ko} & \texttt{ms} & \texttt{ja} & \texttt{km} & \texttt{ru} & \texttt{sv} & \texttt{tr} & \texttt{zh\_cn} & Avg  \\
			\midrule
			\textbf{mLUKE + SP + FiD} & 54.84 & \textbf{30.68} & \textbf{47.41} & \textbf{47.29} & \textbf{33.90} & \textbf{43.13} & \textbf{47.00} & \textbf{43.46} & \textbf{13.34} & \textbf{39.57} & \textbf{29.74} & 24.73 & \textbf{12.14} & \textbf{27.44} & \textbf{18.97} & \textbf{2.57} & \textbf{19.36} & \textbf{28.26} & \textbf{25.52} & \textbf{22.29} & \textbf{21.99}  \\
			(i) mLUKE + FiD & \textbf{54.93} & 29.56 & 46.88 & 45.76 & 33.16 & 42.03 & 46.28 & 42.66 & 13.23 & 38.01 & 29.57 & \textbf{25.36} & 11.45 & 27.28 & 18.37 & 2.53 & 18.59 & 28.22 & 25.43 & 21.98 & 21.67  \\
			(ii) mBERT + FiD & 53.19 & 29.25 & 46.97 & 43.25 & 30.38 & 42.79 & 44.22 & 41.44 & 10.94 & 37.42 & 28.18 & 21.89 & 9.63 & 27.20 & 15.00 & 2.11 & 16.41 & 26.96 & 21.86 & 20.24 & 19.82   \\
			(iii) mBERT + FiE & 49.71 & 29.15 & 42.72 & 41.20 & 30.64 & 40.16 & 38.57 & 38.88 & 8.95 & 33.87 & 25.08 & 21.15 & 6.72 & 24.55 & 15.27 & 6.05 & 15.60 & 25.53 & 20.44 & 13.71 & 18.07   \\
			\bottomrule
	\end{tabular}}
	\caption{Ablation studies on the development sets. mLUKE + SP + FiD is our submission with mLUKE + Sparse-Corroborate-Dense.
		(i) mLUKE + FiD only relies on dense retrieval, and we observe a slight decrease in the F1 score of most languages compared with our submission.
		(ii) mBERT + FiD changes the retriever to mBERT, and we observe a larger drop in F1 score compared to mLUKE in row (i).
		(iii) mBERT + FiE changes Fusion-in-Decoder to Fusion-in-Encoder as in the baseline and we see an even larger drop in F1 score compared with row (ii).}
	\label{table:ablation-overall-dev}
\end{table*}

\paragraph{Ablation Studies}
We conduct ablation studies on our system in Table~\ref{table:ablation-overall-dev}.
We find the biggest gain comes from switching Fusion-in-Encoder (FiE) in the baseline to Fusion-in-Decoder (FiD) from row (iii) to row (ii), even though we did not increase the number of passages for Fusion-in-Decoder and kept it at 20 for the final system.
The second largest gain comes from switching mBERT to mLUKE from row (ii) to row (i).
Finally, the smaller gain comes from switching dense retrieval only to Sparse-Corroborate-Dense, from row (i) to mLUKE + SP + FiD.
We study each of the components in greater detail below.

\paragraph{Entity Representations}

\begin{table}[h]
	\centering
	\resizebox{\linewidth}{!}{
			\begin{tabular}{lcccccccc}
					\toprule
					& \multicolumn{8}{c}{MRR@60} \\
					Model & \texttt{ar} &  \texttt{bn} &  \texttt{fi} &  \texttt{ja} &  \texttt{ko} &  \texttt{ru} &  \texttt{te} & Avg   \\
					\midrule
					mBERT \cite{devlin2019bert} & 0.106 & 0.026 & 0.069 & 0.031 & 0.023 & 0.057 & \textbf{0.050} & 0.362 \\
					mLUKE \cite{ri2021mluke} & 0.106 & \textbf{0.028} & \textbf{0.076} & \textbf{0.035} & \textbf{0.027} & \textbf{0.122} & 0.042 &  \textbf{0.372} \\
					\midrule
					& \multicolumn{8}{c}{Recall@60} \\
					Model & \texttt{ar} &  \texttt{bn} &  \texttt{fi} &  \texttt{ja} &  \texttt{ko} &  \texttt{ru} &  \texttt{te}  & Avg \\
					\midrule
					mBERT \cite{devlin2019bert} & 0.185 & 0.057 & 0.130 & 0.073 & 0.050 & \textbf{0.118} & 0.075 & 0.689 \\
					mLUKE \cite{ri2021mluke} & \textbf{0.189} & \textbf{0.065} & \textbf{0.133} & \textbf{0.078} & \textbf{0.056} & 0.056 & \textbf{0.079} & \textbf{0.723} \\
					\bottomrule
			\end{tabular}}
	\caption{MRR@60 and Recall@60 of passage retrieval for XOR-TyDi QA dev set for different pretrained language models.}
	\label{table:retrieval-recall-lm-choice}
\end{table}

\begin{figure}[htp]
	\centering
	\includegraphics[width=1\linewidth]{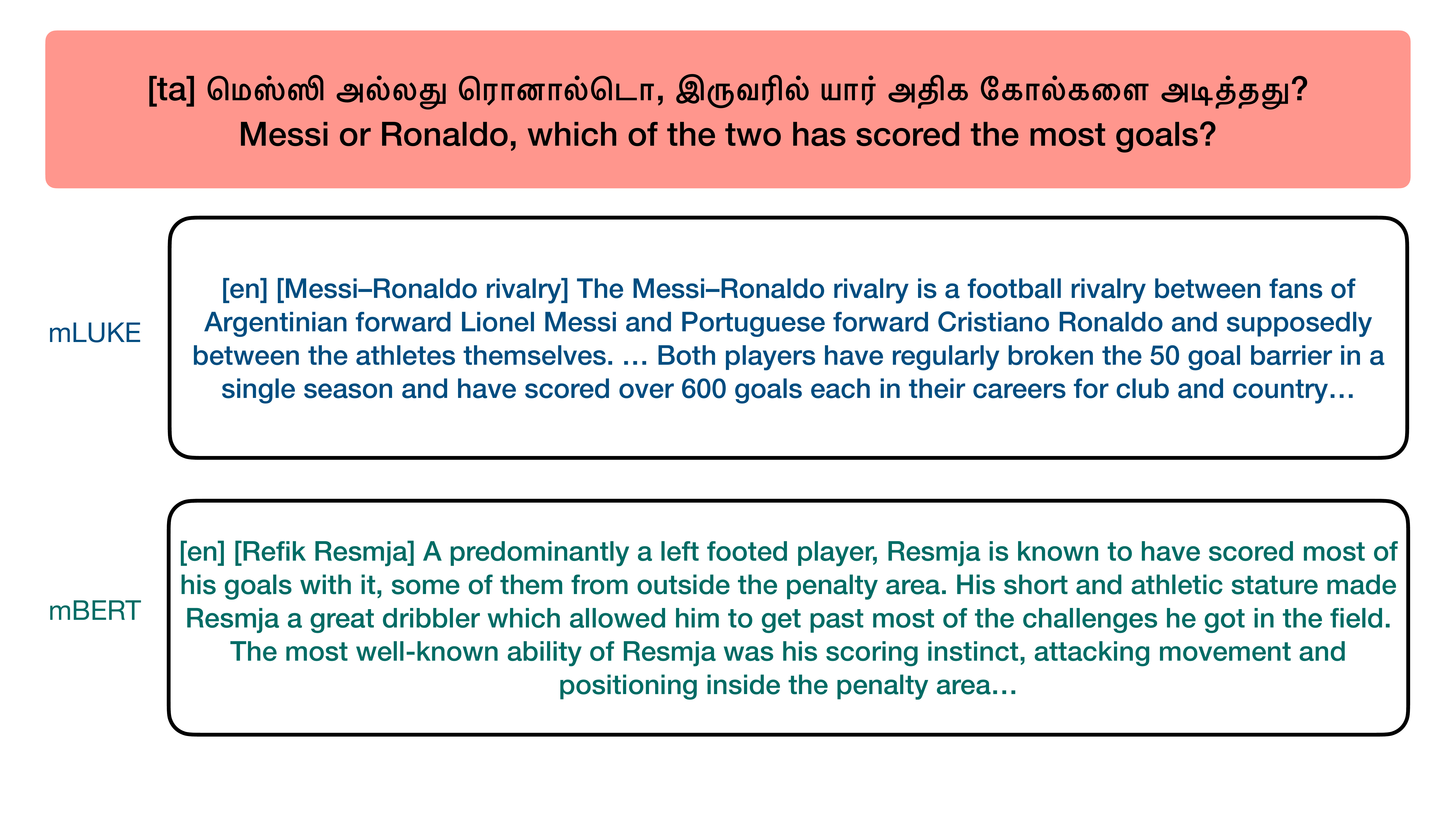}
	\caption{The top passage for a Tamil question retrieved by mBERT and mLUKE.
	We see mLUKE is able to find English passages related to entities Messi and Ronaldo, but mBERT struggles and only finds a general passage related to another unrelated soccer player related to goal scoring.}
	\label{fig:mluke-win}
\end{figure}

To evaluate the passage retrieval component for XOR-TyDi QA, we measure MRR@60 and Recall@60.
We picked 60 because it is the near the maximum number of passages we can feed into Fusion-in-Decoder bound by the GPU memory.
For each question, to determine if a passage is relevant, we use a heuristic.
First, we find the universe set of answers for the questions, which not only contain answers in the same language, but also possibly answers in English using the English answer in the XOR-English Span task \cite{asai2020xor}.
We check if the normalized answer is a substring of the passage text, and if so, we mark the passage as relevant.
Note that this is a proxy for measuring passage relevance, since answers may not necessarily be exact spans / substrings or the same answer may appear as a substring in a non-relevant passage, but we found it to correlate well with end-to-end effectiveness.
We see from Table~\ref{table:retrieval-recall-lm-choice} that overall using mLUKE improves passage retrieval effectiveness.
Qualitatively, we also find examples where the dual encoder trained with mLUKE can find passages cross-lingually with the relevant entity whereas that trained with mBERT could not.
In Figure~\ref{fig:mluke-win}, we see mLUKE can retrieve an English top passage about the soccer players Messi and Ronaldo asked in Tamil, but mBERT returns just an English passage about another soccer player not relevant to the question.

\paragraph{Dense-Sparse Hybrids}

\begin{figure}[htp]
	\centering
	\includegraphics[width=1\linewidth]{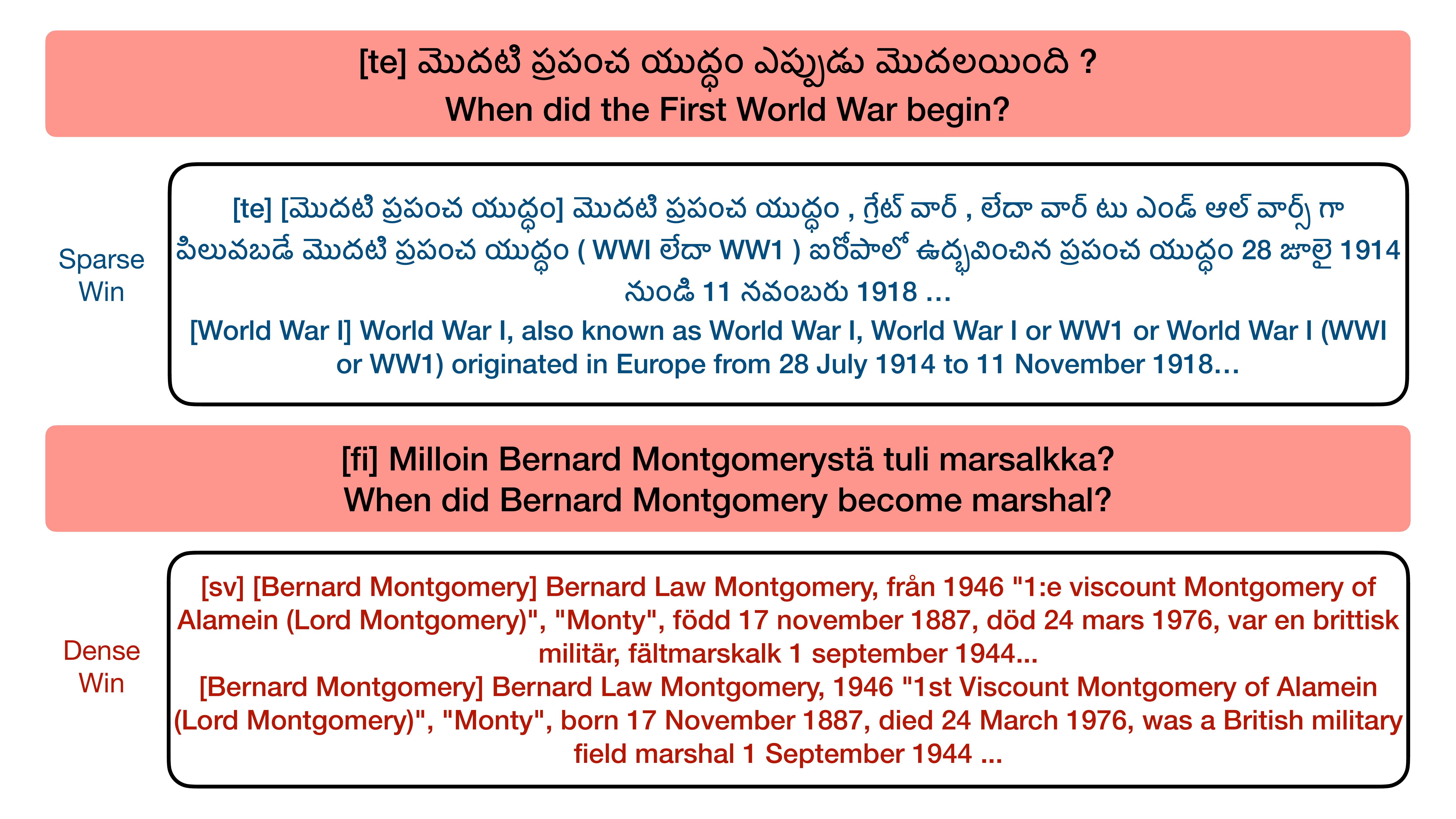}
	\caption{Here we see a highly relevant passage found by sparse monolingual retrieval that is not found by dense retrieval, and a relevant passage found by dense retrieval cross-lingually that is not found by sparse retrieval.}
	\label{fig:sparse-dense-wins}
\end{figure}

Next, we evaluate the benefits of using dense retrieval in conjunction with sparse retrieval as opposed to using only dense retrieval in Table~\ref{table:sparse-dense-hybrid-xor-tydi}.
The dense retriever here is mLUKE.
We see that dense retrieval always works better than sparse retrieval when used independently, and the score combination approach used in Mr. TyDi  \cite{mrtydi} does not outperform dense retrieval in recall, but does improve the MRR.
We use Sparse-Corroborate-Dense, which piggybacks on dense retrieval results, but boosts the ranking of some passages in dense retrieval, and add in additional passages not found by dense retrieval to the end of the top-$K$ list.
Compared to dense only, it is better on both MRR and recall.
When both dense and sparse retrieval finds the same passage, it is a strong signal the passage is relevant.
Nonetheless, sparse retrieval can still find passages that dense retrieval cannot find, and adding these to the candidate passage list passed to the reader can provide additional relevant evidence passages.
In Figure~\ref{fig:sparse-dense-wins}, we see sparse retrieval can find a highly relevant passage related to World War I in Telugu (\texttt{te}) to the Telugu question that cannot be found by dense retrieval, and dense retrieval can find a passage related to Bernard Montgomery cross-lingually in Swedish (\texttt{sv}) to a Finnish (\texttt{fi}) question that cannot be found by sparse retrieval -- they can complement each other.

\begin{table}[h]
	\centering
	\resizebox{\linewidth}{!}{
		\begin{tabular}{lcccccccc}
			\toprule
			& \multicolumn{8}{c}{MRR@60} \\
			Methodology & \texttt{ar} &  \texttt{bn} &  \texttt{fi} &  \texttt{ja} &  \texttt{ko} &  \texttt{ru} &  \texttt{te} & Avg \\
			\midrule
			Sparse Only & 0.088 & 0.023 & 0.640 & 0.024 & 0.018 & 0.051 & 0.032 & 0.299 \\
			Dense Only & 0.106 & 0.028& 0.076 & 0.035 & 0.027 & 0.058 & 0.042 & 0.372 \\
			Combine Score \cite{mrtydi} & 0.113 & 0.029 & 0.076& 0.032 & 0.023 & 0.060 & 0.048 & \textbf{0.382} \\
			Sparse-Corroborate-Dense & 0.110 & 0.029 & 0.074 & 0.032 & 0.026 & 0.063 & 0.049 & \textbf{0.382}\\
			\midrule
			& \multicolumn{8}{c}{Recall@60} \\
			Methodology & \texttt{ar} &  \texttt{bn} &  \texttt{fi} &  \texttt{ja} &  \texttt{ko} &  \texttt{ru} &  \texttt{te} & Avg \\
			\midrule
			Sparse Only & 0.172 & 0.045 & 0.120 & 0.063 & 0.044 & 0.098 & 0.070 & 0.611 \\
			Dense Only & 0.189 & 0.065 & 0.133 & 0.078 & 0.056 & 0.122 & 0.079 & 0.723 \\
			Combine Score \cite{mrtydi} & 0.178 & 0.059 & 0.118 & 0.070 & 0.046 & 0.107 & 0.076 & 0.652 \\
			Sparse-Corroborate-Dense & 0.192 & 0.065 & 0.136 & 0.078 & 0.057 & 0.124 & 0.080 & \textbf{0.733} \\
			\bottomrule
	\end{tabular}}
	\caption{Comparison of various dense-sparse hybrid strategies for original XOR-TyDi QA dev set. The dense retrieval dual encoder used is mLUKE. $max\_frac$ used for Sparse-Corroborate-Dense is 0.2.}
	\label{table:sparse-dense-hybrid-xor-tydi}
\end{table}

\paragraph{Fusion-in-Decoder}

We want to understand the effect of increasing the number of passages sent to the reader by comparing the effectiveness of the reader when there are 20 passages versus 60 passages.
Intuitively, there could be relevant passages found in positions 21-60, which should strengthen the evidence needed to output the final answer.
From Table~\ref{table:reader-context} we observe using more evidence passages consistently improve results, and this scaling advantage is key over Fusion-in-Encoder.
However, due to time limitations, we only used the 20 passages setting for the final shared task submission.

\begin{table}[h]
	\centering
	\resizebox{\linewidth}{!}{
		\begin{tabular}{lcc|cc}
			\toprule
			& \multicolumn{2}{c}{XOR-TyDi QA} & \multicolumn{2}{c}{MKQA} \\
			Number of Passages & EM &  F1 &  EM &  F1  \\
			\midrule
			20 & 31.63 & 38.06 & 16.15 & 20.21 \\
			60 & 33.74 & 41.29 & 17.31 & 21.51 \\
			\bottomrule
	\end{tabular}}
	\caption{Exact Match (EM) and F1 score for different number of passages on the development sets from mLUKE retrieved passages.}
	\label{table:reader-context}
\end{table}

%

\section{Conclusion}

We describe our submission for the MIA 2022 Shared Task and detail some experiments we perform to improve specific components of the system.
We find that using mLUKE \cite{ri2021mluke}, a multilingual language model that models entities during pretraining, combing dense and sparse results using Sparse-Corroborate-Dense, and Fusion-in-Decoder, are helpful for improving the effectiveness for cross-lingual question answering over the baseline.

\section{Acknowledgement}

We thank Yinfei Yang, Wei Wang, and Jinhao Lei for their insightful discussions and feedback on early versions of the paper.

\bibliography{anthology,custom}

\clearpage
 \appendix

 \section{Sparse-Corroborate-Dense Algorithm}
 \label{sec:appendix}

\setcounter{algocf}{0}
\begin{algorithm}[]
	\captionsetup{labelfont={sc,bf}}
	\caption{Sparse-Corroborate-Dense}
	\label{algo:sparse-corroborate-dense}
	\footnotesize
	\SetAlgoLined
	\SetKwInOut{Input}{input}\SetKwInOut{Output}{output}
	\KwData{$DenseHits$, a list of (docid, score) tuples for dense retrieval results;
$SparseHits$, a list of (docid, score) tuples for sparse retrieval results;
$MaxFrac$, a number in $[0, 1]$ representing the max fraction of sparse hits that is allowed to influence the final ranked list;
$K$, the max length of the final ranked list.
	}
	\KwResult{The final ranked list after combining results.}
	\Begin{
		$DenseDocIDToIdx\longleftarrow \{\}$\;
		\For{$idx$, $tup$  $\longleftarrow$ enumerate(DenseHits)}{
			$DenseDocIDToIdx[tup[0]] \longleftarrow idx$\;
		}
		$ReservedSparseSlots \longleftarrow \min(\lfloor MaxFrac \times K \rfloor, len(SparseHits))$ \;
		$FinalHits \longleftarrow []$ \;
		$DocIDsAdded \longleftarrow set()$ \;
		$BackfillSparseHits \longleftarrow []$ \;
		\For(Go through top sparse results, if sparse hit is also in dense, push it to front; else, put it in backfill.){$DocID, SparseScore \longleftarrow SparseHits$}{
			\If{$DocID$ in $DenseDocIDToIdx$}{
				$DenseIdx = DenseDocIDToIdx[DocID]$ \;
				$FinalHits \pluseq DenseHits[DenseIdx]$ \;
				$DocIDsAdded \pluseq DocID$ \;
				$ReservedSparseSlots \minuseq 1$ \;
			} \Else{
				$BackfillSparseHits \pluseq (DocID, SparseScore)$ \;
			}
		}
	
		$i \longleftarrow 0$ \;
		\While(Add rest of dense ids.){$len(FinalHits) < K - ReservedSparseHits$ and $i < len(DenseHits)$}{
			\If{$DenseHits[i]$ not in $DocIDsAdded$}{
				$FinalHits \pluseq DenseHits[i]$ \;
				$DocIDsAdded \pluseq DenseHits[i][0]$ \;
				$i \pluseq 1$ \;
			}
		}
	
		$FinalHits.extend(BackfillSparseHits[:K - len(FinalHits)])$ \;
		\Return{FinalHits}
	}
\end{algorithm}

\end{document}